\documentclass[10pt,twocolumn,letterpaper]{article}

\usepackage{wacv}
\usepackage{times}
\usepackage{epsfig}
\usepackage{graphicx}
\usepackage{amsmath}
\usepackage{amssymb}
\usepackage{booktabs}
\usepackage{wrapfig}
\usepackage{algorithm}
\usepackage{algpseudocode}
\usepackage{algorithmicx} 
\usepackage{color, colortbl}
\usepackage[accsupp]{axessibility}	
\definecolor{Gray}{rgb}{0.9,0.9,1}

\def\wacvPaperID{323} 

\wacvfinalcopy 

\ifwacvfinal
\usepackage[breaklinks=true,bookmarks=false]{hyperref}
\else
\usepackage[pagebackref=true,breaklinks=true,colorlinks,bookmarks=false]{hyperref}
\fi

\begin{document}

\title{Saliency Guided Experience Packing for Replay in Continual Learning}

\author{Gobinda Saha, Kaushik Roy\\
Elmore Family School of Electrical and Computer Engineering\\
Purdue University, West Lafayette, Indiana, USA\\
{\tt\small gsaha@purdue.edu,~kaushik@purdue.edu}
}

\maketitle
\thispagestyle{empty}

\begin{abstract}
Artificial learning systems aspire to mimic human intelligence by continually learning from a stream of tasks without forgetting past knowledge. One way to enable such learning is to store past \textit{experiences} in the form of input examples in episodic memory and \textit{replay} them when learning new tasks. However, performance of such method suffers as the size of the memory becomes smaller. In this paper, we propose a new approach for \textit{experience replay}, where we select the past experiences by looking at the saliency maps which provide visual explanations for the model's decision. Guided by these saliency maps, we pack the memory with only the parts or patches of the input images important for the model's prediction. While learning a new task, we replay these memory patches with appropriate zero-padding to remind the model about its past decisions. We evaluate our algorithm on CIFAR-100, miniImageNet and CUB datasets and report better performance than the state-of-the-art approaches. With qualitative and quantitative analyses we show that our method captures richer summaries of past experiences without any memory increase, and hence performs well with small episodic memory. Codes: \href{https://github.com/sahagobinda/EPR}{\textcolor{magenta}{\texttt{https://github.com/sahagobinda/EPR}}}.
\end{abstract}
\vspace{-15pt}
\section{Introduction}
Recent success in deep learning primarily relies on training powerful models with fixed datasets in stationary environments. However, in the non-stationary setting, where data distribution changes over time, artificial neural networks (ANNs) fail to match the efficiency of human learning. In this setup, humans can learn incrementally leveraging and maintaining past knowledge, whereas ANN training algorithms~\cite{dlbook} overwrite the representations of the past tasks upon exposure to a new task. This leads to rapid performance degradation on the past tasks - a phenomenon known as `Catastrophic Forgetting'~\cite{cat1,cat2}. Continual Learning (CL)~\cite{cl1,cl2} aims to mitigate forgetting while sequentially updating the model on a stream of tasks. 

To overcome catastrophic forgetting, an active line of research in continual learning stores a few training samples from the past tasks as \textit{experiences} in an \textit{episodic memory}.
Variants of \textit{experience replay}~\cite{er,hal,mer,der} have been proposed, where the model is jointly optimized on the samples from both episodic memory and new task. Such methods provide simple yet effective solutions to the catastrophic forgetting especially in online CL~\cite{ocl_surv1} setting where models need to learn from a single pass over the online data stream. However, performance of these methods strongly depends on the size of the episodic memory. The authors in~\cite{ocl} argued that for an optimal performance one needs to store all the past examples in the memory. While experience replay with large memory would yield higher performance, this would essentially mimic the joint optimization process in independent and identically distributed (IID) data setting which puts the effectiveness of CL algorithms into question~\cite{gdumb}. Therefore, recent works~\cite{hal,er} have explored the idea of designing effective experience replay with tiny episodic memory. However, these methods suffer from high forgetting mainly due to overfitting~\cite{overfit} to the small memory samples, thus show suboptimal performance.

In this paper, we propose a continual learning algorithm that trains a fixed-capacity model on an online stream of tasks using a small episodic memory. Our method, referred to as Experience Packing and Replay (\textbf{EPR}), packs the memory with a more informative summary of the past experiences which improves performance of memory replay by reducing overfitting. To this end, we leverage the tools developed in the field of explainable artificial intelligence (XAI)~\cite{xai3,xai2,xai4} that shed light on the internal reasoning process of the ANNs. Among various explainability techniques, saliency methods~\cite{grad-cam,xai1} highlight the part of the input data (image) that the model thinks is important for its final decision. Such analyses reveal that ANNs tend to make predictions based on some localized features or objects belonging to a part of the image whereas the rest of the image appears as background information. Thus we hypothesize that storing and replaying only these important parts of the images would be effective in reminding the networks about the past tasks and hence would reduce forgetting. 

\begin{figure*}[t]
\begin{centering}
  \includegraphics[width=0.85\textwidth,keepaspectratio,page=2]{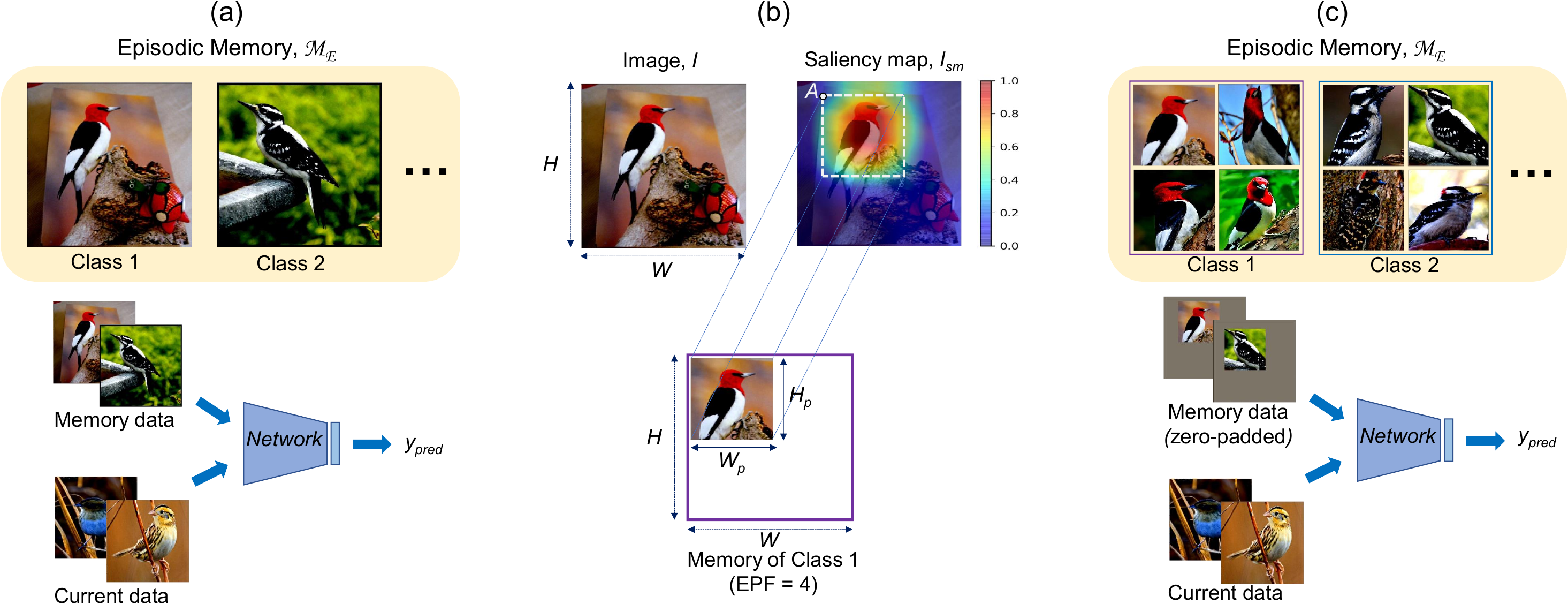}
  \caption{(a) Conventional experience replay method, where full images are stored in the episodic memory and network is jointly trained on these memory data and current task data. (b)-(c) Our experience packing and replay (\textbf{EPR}) method. (b) Experience packing, where only the important part of the image for network prediction is selected using saliency method and stored in the episodic memory with the corner coordinate, $A=(x_{cord},y_{cord})$. (c) Experience packing increases sample diversity per class without memory increase. Stored memory patches are zero-padded and replayed with the current examples.} 
\label{fig:fig1}
\end{centering}
\vspace{-10pt}
\end{figure*}

Therefore, in \textbf{EPR}, after learning each task, instead of storing full images, we identify important patches from different images belonging to each class with saliency method~\cite{grad-cam} and store them in the episodic memory. We introduce \textit{Experience Packing Factor} (EPF) to set the number of patches kept per class and to determine the size of these patches. Thus, with these patches, we create \textit{composite} images (for each class) that have higher diversity and capture richer summaries of past data distributions without increasing the memory size~(Figure~\ref{fig:fig1}). While learning a new task, we retrieve these patches from the memory, zero-pad them to match with the original image sizes, and use them for experience replay. We evaluate our algorithm in standard and directly comparable settings~\cite{er,hal,cope,aser} on image classification datasets including CIFAR-100, miniImageNet, and CUB. We compare EPR with the state-of-the-art (SOTA) methods for varying memory sizes and report better accuracy with least amount of forgetting. We briefly summarize the \textbf{contributions of our work} as follows:
\vspace{-5pt}
\begin{itemize}
    \item We propose a new experience replay method for continual learning - \textbf{EPR}, where we select (and store) episodic memory by identifying the parts of the past inputs important for model's prediction from the saliency maps.~These parts are then replayed with  zero-padding with new data to mitigate catastrophic forgetting.    
    \vspace{-5pt}
    \item We compare EPR with the SOTA methods in both online task-incremental and class-incremental learning~\cite{ocl_surv1} and report significantly better performance.
    \vspace{-5pt}
    \item With comprehensive analyses, we show the effectiveness of saliency methods in selecting the `informative' memory patches which enables EPR to perform well even with tiny episodic memories. 
\end{itemize}
 
 
\vspace{-10pt}
 
\vspace{-5pt}
\section{Related Works}\label{related_works}
Methods for continual learning can be broadly divided into three categories~\cite{cl_survey}.~\textbf{Regularization based methods} penalize changes in important parameters for the past tasks to prevent forgetting. Elastic Weight Consolidation (\textbf{EWC})~\cite{ewc} computes such importance from the Fisher information matrix.
Other works use selective synaptic plasticity~\cite{si,mas}, knowledge distillation~\cite{lwf}, attention map distillation~\cite{lwm}, and variational inference framework~\cite{vcl} for model regularization in continual learning. However, these methods suffer under longer task sequences and perform poorly~\cite{hal} in the online CL setup considered in this paper.  

\textbf{Parameter isolation methods} allocate different subsets of network parameters for each task to overcome forgetting. Some methods~\cite{pgn,den} under this category expand the network for accommodating new tasks, whereas in other methods~\cite{packnet,hat,space} a task-specific sub-network is selected by masking out the parameters. Unlike these methods, we train a fixed-sized network in online CL setting.  

\textbf{Memory based methods} mitigate forgetting by either storing a subset of old examples in the episodic memory for rehearsal~\cite{Robbo,icarl}, or storing important gradient spaces from the past tasks for constrained optimization~\cite{ogd,gpm}, or synthesizing old data from generative models for pseudo-rehearsal~\cite{dgr}. Experience Replay (\textbf{ER})~\cite{Robbo,er} jointly trains the model on the samples from the new tasks and episodic memory.~Several recent methods expand on this idea: Meta-Experience Replay (\textbf{MER})~\cite{mer} combines episodic memory with meta-learning to maximize knowledge transfer and minimize forgetting; \textbf{GSS}~\cite{gss} stores examples in the memory for rehearsal based on the gradients; Maximal Interfered Retrieval (\textbf{MIR})~\cite{mir} selects a minibatch from the episodic memory for replay that incurs maximum change in loss;~Adversarial Shapley value Experience Replay (\textbf{ASER})~\cite{aser} uses shapley value scores for episodic memory selection and retrieval; Hindsight Anchor Learning (\textbf{HAL})~\cite{hal} improves replay by adding an objective term to minimize forgetting on the meta-learned anchor data-points; Dark Experience Replay (\textbf{DER++})~\cite{der} improves ER by replaying network logits along with the ground truth labels of the memory samples. Gradient Episodic Memory (\textbf{GEM})~\cite{gem} and Averaged-GEM (\textbf{A-GEM})~\cite{agem} use samples from the memory to compute gradient constraint so that loss on the past task does not increase. Guo \textit{et al.}~\cite{mega} improved such methods by proposing a loss balancing update rules in \textbf{MEGA}. Ebrahimi \textit{et al.}~\cite{rrr} in \textbf{RRR} stores full images with the corresponding saliency maps in the episodic memory and complements conventional experience replay with a saliency-based regularization objective so that model explanations for the past tasks have minimal drift. Our method, \textbf{EPR}, also uses episodic memory for experience replay. However, unlike these methods, we neither store full images nor store any saliency map in the episodic memory. Rather, leveraging the network's reasoning process, we store only a part (patch) of the image in the memory and use them to remind the model about past decisions in replay. 

\textbf{Online continual learning} (OCL)~\cite{ocl_surv1,ocl_im}  builds on top of continual learning with additional set of realistic yet challenging desiderata.~Here, model observes data one batch at a time and learns from a single pass over these online data stream. Here, model experiences new tasks (consisting of new classes) or data non-stationary (non-IID input distribution) over time. 
In this work, we focus on two widely explored sub-classes of OCL namely online task-incremental learning and online class-incremental learning. In the task-incremental setup, an extra supervisory signal (task-ID) is used to select task-specific head of classes over which inference is performed. Whereas in the class-incremental setup inference is performed without any task-ID.

\vspace{-8pt}
\section{Background and Notations}\label{sec2}
\vspace{-8pt}
\textbf{Continual Learning Protocol.}~For online task-incremental learning we follow the protocol used in~\cite{agem}. Here, a model learns from an ordered sequence of dataset, $\mathcal{D}=\{\mathcal{D}_1, ...~,\mathcal{D}_T\}$ consisting of $T$ tasks, where $\mathcal{D}_k=\{(I_i^k,t_i^k,y_i^k)_{i=1}^{n_k}\}$ is the dataset of the $k$-th task. Each example in these datasets consists of a triplet defined by an input ($I^k \in \mathcal{X}$), an integer task-ID ($t^k \in \mathcal{T}$) and a target vector ($y^k \in \textbf{y}^k$), where $\textbf{y}^k$ is the set of labels specific to task $k$ and $\textbf{y}^k \subset \mathcal{Y}$. As in~\cite{agem,er,hal}, we use the first $K(<T)$ \textit{Cross-Validation} tasks to set the hyperparameters of the continual learning algorithms. The remaining $T-K$ tasks, which is used for training and evaluation, is referred as \textit{Training-Evaluation} tasks. In this setup, the goal is to learn a neural network,  $f_{\theta}:\mathcal{X}\times\mathcal{T} \rightarrow \mathcal{Y}$, parameterized by $\theta \in \mathbb{R}^P$, that maps any input pair $(I,t)$ to its target output $y$ and maintains performance on all the prior tasks. For online class-incremental learning, task-ID ($t^k$) is omitted.  

\textbf{Saliency Map Generation.} Saliency methods provide visual explanations for the model predictions in terms of relevant features in the input. For example, for an input RGB image, $I \in \mathbb{R}^{W\times H\times C}$ belonging to class $c$, these methods generate a \textit{saliency map}, $I_{sm} \in \mathbb{R}^{W\times H}$ by assigning high intensity values to the relevant image regions that contribute to the model decision. The saliency map is generated by: 
\begin{equation}\label{eq1}
     I_{sm}=\texttt{XAI}~(f_\theta,I,c)
\end{equation}
where $\texttt{XAI}$ is a saliency method. Simonyan~\textit{et al.}~\cite{xai3} first generated saliency maps using a pre-trained neural network. Later works improved the quality of saliency maps~\cite{sal1,sal2} and reduced the cost for saliency computation~\cite{xai2,grad-cam}. In this work, we primarily use Gradient-weighted Class Activation Mapping (\textbf{Grad-CAM})~\cite{grad-cam} as the saliency method. It generates class-specific saliency maps based on gradients back-propagated to the later convolutional layers, given the model prediction. We describe the steps in detail in Appendix~\ref{App_saliency}. We also analyze the impact of other saliency methods such as Grad-CAM++~\cite{grad_cam_pp}, Smooth-Grad~\cite{smoothgrad} and FullGrad~\cite{fullgrad} on EPR performance in Section~\ref{sec:res}. 

\vspace{-8pt}
\section{Experience Packing and Replay (EPR)}\label{method_epr}
\vspace{-5pt}
\textbf{Experience Replay.} Continual learning algorithms, especially OCL methods, have achieved SOTA performance using~\textit{experience replay}~\cite{stream,der,hal}.~These methods update the model, $f_\theta$ while storing a few samples from the training data (either randomly~\cite{er,agem,mer} or selectively~\cite{mir,gss,aser}) into a replay buffer called episodic memory, $\mathcal{M}_E$. When data from a new task becomes available, the model is jointly trained on both the current and the episodic memory examples (Figure~\ref{fig:fig1}(a)). Thus, experience replay from $\mathcal{M}_E$ mitigates catastrophic forgetting by reminding the network about the prior tasks. However, performance of these methods shows strong dependence on the number of samples kept in the memory. Though with larger $\mathcal{M}_E$ replay yields better performance, designing effective experience replay with small episodic memory~\cite{er,hal} still remains an open research problem. This is because, the model performance becomes highly sensitive to the examples stored in a smaller-sized $\mathcal{M}_E$. Moreover, lack of sample diversity (per class) leads to higher overfitting to the memory examples, which causes loss of generalization for the past tasks leading to catastrophic forgetting~\cite{overfit}. To overcome these issues, we propose a method to select and store only patches of images, instead of full images, from the past tasks in $\mathcal{M}_E$. This enables us to pack diverse experiences from an image class without any memory increase. Next, we introduce the concept of \textit{Experience Packing Factor} and describe how we select these patches and use them during replay.     

\textbf{Experience Packing Factor (EPF).} Let's consider $n_{sc}$ to be the number of (episodic) memory slots assigned for each class. Here, one memory slot can contain one full training image. For a given image, $I \in \mathbb{R}^{W\times H\times C}$ and the target patch size (from this image) $I_p \in \mathbb{R}^{W_p\times H_p\times C}$ (with $W_p\leq W$ and $H_p\leq H$) Experience Packing Factor (EPF) is defined as the following ratio:
\begin{equation}
    \textnormal{EPF} = n_{sc} \frac{W\times H}{W_p\times H_p}.
\end{equation}
EPF is integer-valued and it refers to the number of patches one can fit into the given memory slot, $n_{sc}$, for any particular class. In our design, we consider square images ($W=H$) and patches ($W_p=H_p$) and set EPF as a hyperparameter. Thus, for a given EPF we determine the image patch size as:
\begin{equation}\label{epf2}
    W_p = \sqrt{\frac{n_{sc}}{\textnormal{EPF}}} W.
\end{equation}
We take the floored integer value of $W_p$. Equation~\ref{epf2} tells us, for instance, to pack $4$ patches (EPF$=4$) into $1$ memory slot ($n_{sc}=1$), the patch width (height) should be half of the full image width (height).

\textbf{Memory Patch Selection and Storage.} Explainability techniques~\cite{xai2,grad-cam,xai1} reveal that ANN bases its decision on the class-discriminative localized features in the input data (image).~Hence, we propose to store only the important part (patch) of that image and use it during replay to remind the network about its past decision. We identify these patches from the saliency maps (Section~\ref{sec2}) of the full images. Therefore, while learning each task, we store only a few training images in a small, fixed-sized FIFO `temporary ring buffer', $\mathcal{M}_T$~\cite{er}. At the end of each task, we extract the desired number of patches from these images using saliency maps and add them to the memory, $\mathcal{M}_E$. Note that, images from $\mathcal{M}_T$ are only used for patch selection and not used in experience replay in the later tasks. Once the memory patches are stored in $\mathcal{M}_E$, the temporary buffer, $\mathcal{M}_T$ is freed up to be reused in the next task. If we assume that data from $k$-th task is available to the model until it sees the next tasks (as in~\cite{rrr}), $\mathcal{M}_T$ is not needed.     

Let $f_\theta^k$ is the trained model after task $k$. For each examples in $\mathcal{M}_T$ : $(I,t^k,c) \sim \mathcal{M}_T$ we generate the corresponding saliency map, $I_{sm}$ using Equation~\ref{eq1}. For given $n_{sc}$ and chosen EPF, we obtain the (square) patch size ($W_p\times W_p$). Then, we \texttt{average-pool} the saliency map, $I_{sm}$ with kernel size $W_p\times W_p$ and stride (hyperparameter), $S_{sm}$. We store the top left coordinate ($x_{cord}, y_{cord}$) of the kernel (patch) that corresponds to maximum average-pooled value. In other words, we identify a square region (of size $W_p\times W_p$) in the saliency map that has the maximum average intensity (Figure~\ref{fig:fig1}(b)). We obtain the memory patch from the image, $I$ by :
\begin{equation}\label{eq:epr_eq4}
    I_p = I(x_{cord}:x_{cord}+W_p, y_{cord}:y_{cord}+W_p ).
\end{equation}
In our design, we keep a few more image samples in $\mathcal{M}_T$ per class than the number of image patches we store in $\mathcal{M}_E$. As we will be using these patches with zero-padding (discussed next) for replay, for storage in $\mathcal{M}_E$, we want to prioritize the patches that after zero-padding gives (or remain closer to) the correct class prediction. Thus we \texttt{Zero-pad} each image patch, $I_p$ and check the model prediction. At first, we populate the memory with the patches for which model gives correct prediction. Then we fill up the remaining slots in $\mathcal{M}_E$ by the patches for which correct class is in model's \textit{Top3} predictions. Any remaining memory slot is filled up from the remaining patches irrespective of model predictions. Each selected image patch is then added to $\mathcal{M}_E = \mathcal{M}_E\cup\{(I_p,t^k,c,x_{cord},y_{cord})\}$, with task-ID, class label and localizable coordinates in the original image.   
\setlength{\textfloatsep}{7pt}
\begin{algorithm}[!t]
  \footnotesize
  \caption{Experience Packing and Replay (EPR)}\label{epr_algo}
  \begin{algorithmic}[1]
    \Statex \textbf{Inputs:} {$T$: No. of tasks; $n_{sc}$: No. of memory slots per class; EPF: No. of patches per class; $W$: image width (height); $\alpha$: learning rate; $\mathcal{D}^{train}$: training dataset; $f_\theta$: model with param $\theta$}; $n$: mini-batch size
    \Statex \textbf{Output:} Updated model, $f_\theta$
    \State $\mathcal{M}_T\gets \{\}$ \Comment{`temporary ring' buffer}
    \State $\mathcal{M}_E\gets \{\}$ \Comment{episodic memory}
    \State $W_p\gets f_{patch}(n_{sc},\textnormal{EPF},W)$ \Comment{patch size from Eq.~\ref{epf2}}
    \For{$t \in {1,2,.....,T}$} 
        \For {$\mathcal{B}_t\overset{\mathrm{n}}{\sim} \mathcal{D}_t^{train}$ } \Comment{from current dataset}
        \State $(\mathcal{B}_{\mathcal{M}_E},\textbf{x}_{cord},\textbf{y}_{cord})\overset{\mathrm{n}}{\sim} \mathcal{M}_E$ \Comment{from episodic memory}
        \State $\mathcal{B}_{\mathcal{M}_E}\gets\texttt{Zero-pad}(\mathcal{B}_{\mathcal{M}_E},\textbf{x}_{cord},\textbf{y}_{cord})$ 
        \State $\theta\gets \textnormal{SGD}(\theta,\mathcal{B}_t\cup\mathcal{B}_{\mathcal{M}_E},\alpha)$ \Comment{update model with replay} 
        \State $\mathcal{M}_T\gets \texttt{Update-Ring-Buffer}(\mathcal{M}_T,\mathcal{B}_t)$ 
        \Comment{\cite{er}}
        
        \EndFor
        \State $\mathcal{M}_E\gets  \texttt{UPDATEMEMORY}(\mathcal{M}_E,\mathcal{M}_T,f_\theta,\textnormal{EPF},W_p)$ \Comment{Appendix \ref{App_algorithm} }
        \State $\mathcal{M}_T\gets \{\}$ \Comment{clear `temporary ring' buffer}
     \EndFor
  \end{algorithmic}
\end{algorithm}

\textbf{Replay with Memory Patches.} Since the patches stored in $\mathcal{M}_E$ are smaller in size than the original images, we \texttt{Zero-pad} these patches (Figure~\ref{fig:fig1}(c)) each time we use them for experience replay. While zero-padding we place these patches in the `exact' position of their original images using the coordinate values ($x_{cord}, y_{cord}$). Each sample,   
\begin{equation}
    I_{rep} = \texttt{Zero-pad}~(I_p, x_{cord}, y_{cord}) 
\end{equation}
for replay will thus have the same dimensions as the samples of the current task. Throughout the paper, we use zero-padding with the exact placement of the memory patches for replay unless otherwise stated. We discuss other choices for memory patch padding and placement in Section~\ref{sec:res}. The pseudo-code of our algorithm is given in Algorithm~\ref{epr_algo}. 

\section{Experimental Setup}\label{exp_setup}
\vspace{-8pt}
Here, we describe the task-incremental learning setup. In Section~\ref{sec:res} we discuss the class-incremental learning setup.

\textbf{Datasets.} We evaluate our algorithm on three image classification benchmarks widely used in continual learning.~\textbf{Split CIFAR}~\cite{gem} consists of splitting the original CIFAR-100 dataset~\cite{cifar} into 20 disjoint subsets, each of which is considered as a separate task containing 5 classes. \textbf{Split miniImageNet}~\cite{er,hal,acl} is constructed by splitting 100 classes of miniImageNet~\cite{miniI} into 20 tasks where each task has 5 classes.~\textbf{Split CUB}~\cite{agem,er,mega} is constructed by splitting 200 bird categories from CUB dataset~\cite{cub} into 20 tasks where each task has 10 classes. The dataset statistics are given in Appendix~\ref{App_dataset}. We do not use any data augmentation. All datasets have 20 tasks ($T=20$), where first 3 tasks ($K=3$) are used for hyperparameter selection while the remaining $17$ tasks are used for training. We report performances on the held-out test sets from these 17 tasks.     

\textbf{Network Architectures and Training.} For CIFAR and miniImageNet, we use a reduced ResNet18~\cite{hal} with three times fewer feature maps across all layers. For CUB, we use a standard ImageNet pretrained ResNet18~\cite{agem,er}. Similar to~\cite{agem,er,hal,mega}, we train and evaluate our algorithm in `multi-head' setting~\cite{eval1} where a task-ID is used to select a task-specific classifier. All the models are trained using SGD with batch size of $n=10$ for both the current and memory examples. All experiments are averaged over $5$ runs using different random seeds, where each seed corresponds to a different model initialization and dataset ordering among tasks. A list of hyperparameters along with the EPFs used in these experiments is given in Appendix~\ref{App_hyplist}. 

\textbf{Baselines.} From memory based methods, we compare with A-GEM~\cite{agem}, MIR~\cite{mir}, MER~\cite{mer}, MEGA-I~\cite{mega}, DER++~\cite{der}, ASER~\cite{aser}, HAL~\cite{hal} and experience replay~\cite{er} with ring (ER-RING) and reservoir (ER-Reservoir) buffer. We also compare with EWC~\cite{ewc} which uses regularization and RRR~\cite{rrr} that uses both memory replay and regularization. We include two non-continual learning baselines: Finetune and Multitask. Finetune, where a single model is trained continually without any memory or regularization, gives performance lower bound. Multitask is an oracle baseline where a model is trained jointly on all tasks.         


\textbf{Performance Metrics.} We evaluate the classification performance using the \textbf{ACC} metric, which is the average test classification accuracy of all tasks. We report backward transfer, \textbf{BWT} to measure the influence of new learning on the past knowledge. For instance, negative BWT indicates forgetting. Formally, ACC and BWT are defined as:
\begin{equation}
    \textnormal{ACC} = \frac{1}{T}\sum_{i=1}^T R_{T,i}, \;\;
    \textnormal{BWT} = \frac{1}{T-1}\sum_{i=1}^{T-1} - R_{i}^T
\end{equation}
Here, $T$ is the total number of sequential tasks, $R_{T,i}$ is the accuracy of the model on $i^{th}$ task after learning the $T^{th}$ task sequentially~\cite{gem}, and $R_{i}^T=\underset{l \in \{1,...,T-1\}}{\textnormal{max}}(R_{l,i} - R_{T,i})$~\cite{hal}.

\renewcommand{\arraystretch}{1.1}
\begin{table*}[t]
\centering
\caption{Performance comparison in task-incremental learning setup.~(*) indicates results for CIFAR and miniImageNet are reported from HAL~\cite{hal} and results for CUB are reported from ER-RING~\cite{er}.~($\dagger$) indicates results are reported from ER-RING. We (re) produced all the other results. Average and standard deviations are computed over $5$ runs for different random seeds. No. of memory slots per class, $n_{sc}$=$\{1,2\}$ refers to memory size, $| \mathcal{M}_E |$=$\{85,170\}$ for CIFAR and miniImageNet, and $| \mathcal{M}_E |$=$\{170,340\}$ for CUB.}
\vspace{-8pt}
\scalebox{0.78}{
\begin{tabular}[t]{@{}clccccccccc@{}}
\toprule
&\multicolumn{1}{c}{}
&\multicolumn{2}{c}{\textbf{Split CIFAR}} 
&\multicolumn{3}{c}{\textbf{Split miniImageNet}} 
&\multicolumn{3}{c}{\textbf{Split CUB}} \\

\midrule
\textbf{$n_{sc}$}&\textbf{Methods}& {ACC (\%)}  & {BWT}  && {ACC (\%)}  & {BWT} && {ACC (\%)}  & {BWT}\\ \midrule
-&Finetune*    &  42.9 $\pm$ 2.07 & - 0.25 $\pm$ 0.03 &&  34.7 $\pm$ 2.69 & - 0.26 $\pm$ 0.03  && 55.7 $\pm$ 2.22 & - 0.13 $\pm$ 0.03 \\
&EWC* ~\cite{ewc}   &   42.4 $\pm$ 3.02 & - 0.26 $\pm$ 0.02  &&   37.7 $\pm$ 3.29 & - 0.21 $\pm$ 0.03 &&  55.0 $\pm$ 2.34 & - 0.14 $\pm$ 0.02 \\
\midrule
1&RRR~\cite{rrr} &   - & -  && -  & -  &&  62.9 $\pm$ 1.33 & - 0.04 $\pm$ 0.01\\
&A-GEM*~\cite{agem}  &   54.9 $\pm$ 2.92 & - 0.14 $\pm$ 0.03  &&  48.2 $\pm$ 2.49 & - 0.13 $\pm$ 0.02 && 62.1 $\pm$ 1.28 & - 0.09 $\pm$ 0.01\\
&MIR*~\cite{mir}    &   57.1 $\pm$ 1.81& - 0.12 $\pm$ 0.01  && 49.3 $\pm$ 2.15 & - 0.12 $\pm$ 0.01 && -   & - \\
&MER*~\cite{mer}    &   49.7 $\pm$ 2.97& - 0.19 $\pm$ 0.03  && 45.5 $\pm$ 1.49 & - 0.15 $\pm$ 0.01 && 55.4 $\pm$ 1.03  & - 0.10 $\pm$ 0.01\\
&MEGA-I~\cite{mega} & 55.2 $\pm$ 1.21 & - 0.14 $\pm$ 0.02  && 48.6 $\pm$ 1.11  & - 0.10 $\pm$ 0.01 && 65.1 $\pm$ 1.30 & - 0.05 $\pm$ 0.01\\
&DER++~\cite{der} &  54.0 $\pm$ 1.18 & - 0.15 $\pm$ 0.02 &&  48.3 $\pm$ 1.44& - 0.11 $\pm$ 0.01 &&  66.8 $\pm$ 1.36 & - 0.04 $\pm$ 0.01\\
&ASER~\cite{aser} &  55.4 $\pm$ 1.17 & - 0.16 $\pm$ 0.01 && 48.2 $\pm$ 1.43 & - 0.09 $\pm$ 0.01 &&  66.2 $\pm$ 1.63 & - 0.07 $\pm$ 0.02\\
&ER-Reservoir$\dagger$~\cite{er} &  53.1 $\pm$ 2.66 & - 0.19 $\pm$ 0.02 &&  44.4 $\pm$ 3.22& - 0.17 $\pm$ 0.02 &&  61.7 $\pm$ 0.62 & - 0.09 $\pm$ 0.01\\
&ER-RING*~\cite{er} &  56.2 $\pm$ 1.93 & - 0.13 $\pm$ 0.01 &&  49.0 $\pm$ 2.61& - 0.12 $\pm$ 0.02 &&  65.0 $\pm$ 0.96 & - 0.03 $\pm$ 0.01\\
\rowcolor{Gray}
&\textbf{EPR (Ours)} & \textbf{58.5 $\pm$ 1.23}& - \textbf{0.10 $\pm$ 0.01}  && \textbf{51.9 $\pm$ 1.57}  & - \textbf{0.06 $\pm$ 0.01} && \textbf{72.1 $\pm$ 0.93} & - \textbf{0.02 $\pm$ 0.01}\\
\midrule
2&RRR &   - & -  && -  & -  &&  67.1 $\pm$ 1.27 & - 0.03 $\pm$ 0.01\\
&MEGA-I & 57.6 $\pm$ 0.87 & - 0.12 $\pm$ 0.01  && 50.3 $\pm$ 1.14  & - 0.08 $\pm$ 0.01 && 67.8 $\pm$ 1.30 & - 0.04 $\pm$ 0.01\\
&DER++ &  56.3 $\pm$ 0.98 & - 0.14 $\pm$ 0.01 &&  50.1 $\pm$ 1.14& - 0.09 $\pm$ 0.01 &&  70.7 $\pm$ 0.62 & - 0.03 $\pm$ 0.01\\
&ASER &  57.5 $\pm$ 1.21 & - 0.13 $\pm$ 0.01 &&  50.1 $\pm$ 1.07 & - 0.08 $\pm$ 0.01 &&  69.9 $\pm$ 0.85 & - 0.05 $\pm$ 0.01\\
&ER-RING & 58.6 $\pm$ 2.68 & - 0.12 $\pm$ 0.01 &&  51.2 $\pm$ 2.06& - 0.10 $\pm$ 0.01 &&  68.3 $\pm$ 1.13 & - 0.02 $\pm$ 0.01\\
&HAL*~\cite{hal} &   60.4 $\pm$ 0.54& - 0.10 $\pm$ 0.01  && 51.6 $\pm$ 2.02 & - 0.10 $\pm$ 0.01 && -  & - \\
\rowcolor{Gray}
&\textbf{EPR (Ours)} & \textbf{60.8 $\pm$ 0.35}  & - \textbf{0.09 $\pm$ 0.01}  && \textbf{53.2 $\pm$ 1.45} & - \textbf{0.05 $\pm$ 0.01}&& \textbf{73.5 $\pm$ 1.30} & - \textbf{0.01 $\pm$ 0.01}\\
\midrule
-&MultiTask* & 68.3 & - && 63.5 & - && 65.6 & -\\
\bottomrule
\end{tabular}
}
\label{table1}
\vspace{0pt}
\end{table*}

\renewcommand{\arraystretch}{1.3}
\begin{table*}[t]
\caption{(a) ACC(\%) comparison in class-incremental learning setup, where results for CIFAR-100 and miniImageNet baselines are taken from~\cite{cope} and \cite{aser} respectively. (b) Impact of padding, placement and selection method of memory patches on EPR performance ($n_{sc}=1$).}
\vspace{-5pt}
\begin{minipage}[t]{0.1\linewidth}
\centering
\scalebox{0.69}{
    \begin{tabular}[t]{@{}lcccc@{}}
        &  \multicolumn{4}{c}{\textbf{(a)}} \\
        \toprule
        &  \multicolumn{2}{c}{\textbf{CIFAR-100} (20 Tasks)} &  \multicolumn{2}{c}{\textbf{miniImageNet} (10 Tasks)}\\
        \midrule
        \textbf{Methods}& {$|\mathcal{M}_E|$=1k}  & {$|\mathcal{M}_E|$=2k}  & {$|\mathcal{M}_E|$=1k}  & {$|\mathcal{M}_E|$=2k} \\ 
        \midrule
        GSS~\cite{gss}  & 7.6 $\pm$ 1.81 &  9.9 $\pm$ 1.17  & 7.5 $\pm$ 0.50  &  10.7 $\pm$ 0.80 \\
        MIR~\cite{mir}  & 9.0 $\pm$ 1.20 & 12.0 $\pm$ 1.84  & 8.1 $\pm$ 0.30  & 11.2 $\pm$ 0.70 \\
        ER~\cite{er}  & 7.9 $\pm$ 1.98 & 11.9 $\pm$ 3.42  & 8.7 $\pm$ 0.40  & 11.8 $\pm$ 0.90 \\
        ASER~\cite{aser}  & -  & -   & 12.2 $\pm$ 0.80  &  14.8 $\pm$ 1.10 \\
        CoPE~\cite{cope}  & 10.7 $\pm$ 1.13 & 14.8 $\pm$ 1.18  & -  & -  \\
        \rowcolor{Gray}
        \textbf{EPR (Ours)}  & \textbf{13.7 $\pm$ 0.84} &  \textbf{16.3 $\pm$ 0.86}  & \textbf{13.8 $\pm$ 0.23}  & \textbf{15.6 $\pm$ 0.31}\\
        \bottomrule
    \end{tabular}
}
\end{minipage}
\hspace{0.5cm}
\begin{minipage}[t]{1.15\linewidth}
\centering
\scalebox{0.73}{
\begin{tabular}[t]{@{}lcccccc@{}}
&  \multicolumn{4}{c}{\textbf{(b)}} \\
\toprule
&  \multicolumn{2}{c}{\textbf{CIFAR}} & \multicolumn{2}{c}{\textbf{miniImageNet}} & \multicolumn{2}{c}{\textbf{CUB}}\\
\midrule
\textbf{Methods}& {ACC (\%)}  & {BWT}  & {ACC (\%)}  & {BWT} & {ACC (\%)}  & {BWT}\\ \midrule
EPR (\texttt{Zero-pad,exact}) & \textbf{58.5} & - \textbf{0.10}  & \textbf{51.9}  & - \textbf{0.06} & \textbf{72.1} & - \textbf{0.02}\\
EPR (\texttt{Zero-pad,random}) & 57.0 & - 0.11  & 51.5  & - \textbf{0.06} & 71.9 & - \textbf{0.02 }\\

EPR (\texttt{Random-pad,exact}) & 57.2 & - 0.11  & 49.7  & - 0.07  & 71.5 & - \textbf{0.02 }\\
\midrule
Random Snip \& Replay & 53.6  & - 0.14  & 49.5  & - 0.08  & 67.4  & - 0.05 \\
Input Compression & 56.7  & - 0.12  & 49.6   & - 0.08  & 69.6  & - 0.03 \\
\bottomrule
\end{tabular}
}
\end{minipage}
\vspace{-5mm}
\label{table2}
\end{table*}

\section{Results and Analyses}\label{sec:res}
\vspace{-5pt}
\textbf{Task-incremental Learning Performance.} First, we compare the performance~(ACC and BWT) of EPR with the baselines. Table~\ref{table1} summarizes the results, where for a given $n_{sc}$, episodic memory (of either ring or reservoir type) can store up to $| \mathcal{M}_E |$ examples. Here, $n_{sc}$ is the number of memory slots per class and the memory size, $| \mathcal{M}_E |$ is :
\begin{equation}
    |\mathcal{M}_E| = n_{sc}\times \textnormal{no. of classes per task}\times \textnormal{no. of tasks}.
\end{equation}
Results in Table~\ref{table1} show that performance of EWC is almost identical to the `Finetune' baseline. This indicates that such method is ill-suited for online CL setup. When one memory slot is assigned per class ($n_{sc}=1$), our method (EPR) outperforms A-GEM and MEGA-I considerably for all the datasets. Moreover, compared to the other experience replay methods, such as MIR, MER, DER++, and ER, EPR achieves $\sim2\%$ and $\sim3\%$ accuracy improvement for CIFAR and miniImageNet respectively with least forgetting. In CUB, EPR obtains $\sim5\%$ accuracy improvement over these baselines with only $\sim2\%$ forgetting. 
For all datasets, EPR considerably outperforms ASER which shows our saliency based memory storage offers better solution for small-memory experience replay compared to the shapely value based memory selection in ASER. Moreover, for CUB dataset, EPR obtains $\sim9\%$ better accuracy with $\sim2\%$ less forgetting compared to RRR. This demonstrate the benefit of saliency based input selection (for replay) in EPR over the saliency map regularization in RRR.

Finally, we compare EPR with HAL~\cite{hal} which holds the SOTA performance in this setup. For the miniImageNet tasks, EPR (with $n_{sc}=1$) achieves slightly better accuracy than HAL, whereas HAL outperforms EPR at the CIFAR tasks. However, in addition to the ring buffer, HAL uses extra memory to store anchor points having the same size of the full images for each class. Thus effectively, HAL uses two memory slots per class ($n_{sc}=2$). In Table~\ref{table1}, we compare EPR with HAL where EPR also uses two memory slots per class. Under this iso-episodic memory condition, EPR has better accuracy and lower forgetting than HAL for both datasets. In this case ($n_{sc}=2$), EPR outperforms all the other methods significantly. For all datasets, amount of forgetting in EPR reduces with increase in memory size.       

\textbf{Class-incremental Learning Performance.} In Table~\ref{table2}(a), we compare EPR with SOTA baselines in class-incremental learning setup for varying episodic memory sizes. For CIFAR-100 (20 Tasks) and miniImageNet (10 Tasks) experiments, we used training setup the similar to Continual Prototype Evolution (CoPE)~\cite{cope} and ASER~\cite{aser} respectively. In this setup, `single-head' inference is performed without task-ID and EPR outperforms all the baselines achieving up to $3\%$ ACC gain. All the subsequent analyses are performed in task-incremental setup.       

\textbf{Padding and Placement of Memory Patches.} Next, we analyze the impact of different types of padding and placement of the memory patches on the EPR performance.~For padding we have two different choices: we can either \texttt{Zero-pad} these patches or we can pad these patches with pixels sampled from normal Gaussian distribution, which we refer to as \texttt{Random-pad}. Similarly, we can place these patches either in the \texttt{exact} position of their original image using stored coordinate values or we can place them at \texttt{random} positions. Table~\ref{table2}(b) shows that across all datasets, \texttt{exact} placement works slightly better than \texttt{random} placement. This indicate that neural network remembers the past tasks better if it finds the class-discriminative features in their original position during replay. For all the datasets, zero-padding performs better than random-padding (Table~\ref{table2}(b)), which indicates that removing the background information completely serves as a better reminder of past tasks for the network. Thus, in all our experiments we use zero-padding with exact placement. For this, we store a 2D coordinate value per memory patch which has an insignificant overhead compared to $| \mathcal{M}_E |$ .   

\begin{figure*}[!t]
\begin{centering}
  \includegraphics[width=0.80\textwidth,keepaspectratio,page=4]{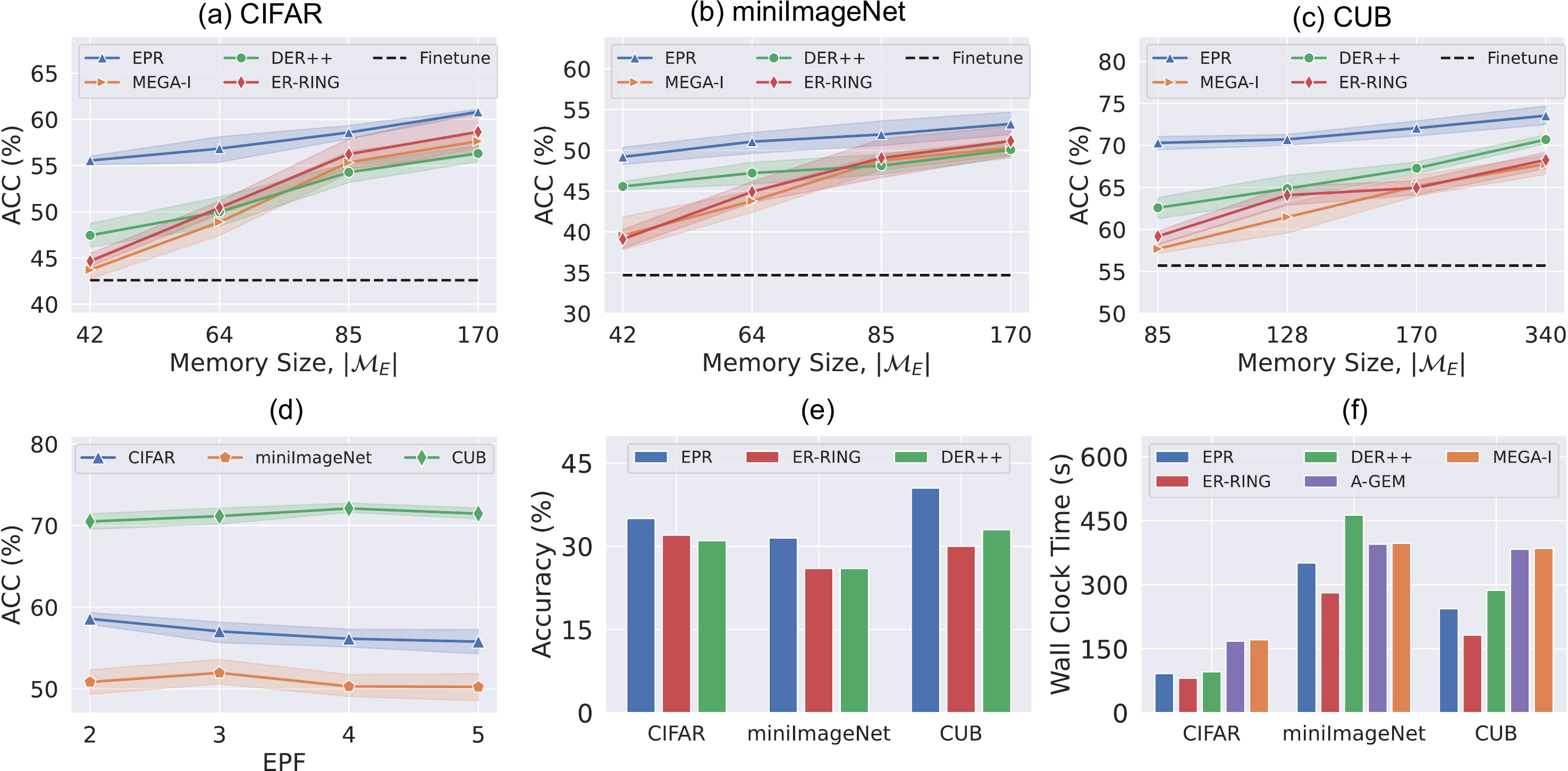}
  \caption{Comparison of ACC for varying memory sizes for (a) Split CIFAR, (a) Split miniImageNet, and (c) Split CUB dataset. (d) ACC for different Experience Packing Factors (EPFs) for different datasets in EPR (for $n_{sc}=1$). (e) Joint training accuracies on episodic memory data compare buffer informativeness (for $n_{sc}=1$). (f) Total wall-clock training time for learning all the tasks sequentially.}  
\label{fig:result_figs}
\end{centering}
\vspace{-12pt}
\end{figure*}

\textbf{Effectiveness of Saliency Guided Memory Selection.} A simple alternative to our saliency guided memory patch selection is to randomly select a patch (of size $W_p\times W_p$) from the original image and use it for replay with zero-padding. We refer to this method as `Random Snip \& Replay' and compare it with EPR in Table~\ref{table2}(b). For CIFAR and CUB, EPR achieves~$\sim5\%$ and for miniImageNet EPR achieves~$\sim2.5\%$ better accuracy than this baseline. We investigate another simple `Input Compression' baseline, where we down-sample the images to desired sizes (comparable to EPR) and store in $\mathcal{M}_{E}$ and for replay up-sample them to the original sizes.~Results in Table.~\ref{table2}(b) shows, EPR outperforms this method by up to ~$\sim2.5\%$ implying that input compression leads to higher loss in information. These results show that saliency based memory patch selection plays a key role in enabling high performance in EPR.

\textbf{Experience Replay with Tiny Episodic Memories.} Next, we study the impact of buffer size, $| \mathcal{M}_E |$ on the performance of experience replay methods. In Table~\ref{table1}, we reported the results for $n_{sc}=\{1,2\}$ to provide a direct comparison to the SOTA works. Here, we analyze whether it is possible to reduce the memory size further and still have an effective experience replay. This means, we consider the fractional values for $n_{sc}$. In such cases, for instance, $n_{sc}=0.5$ means only half of the seen classes will have a sample stored in $\mathcal{M}_E$. Understandably this is a challenging condition for standard experience replay as many classes will not have any representation in the memory, leading to a sharp drop in performance. However, in our method, we can set an appropriate EPF ($\geq1$) for any given $n_{sc}(>0)$ and use Equation~\ref{epf2} to get the size of the memory patches.~This allows us to pack representative memory patches from each class and preserve performance of experience replay. In Figure~\ref{fig:result_figs}(a)-(c) we show how the performance (ACC) of different memory replay methods varies with the memory size for different datasets. Here we consider, $n_{sc}=\{0.5,0.75,1,2\}$ which corresponds to $| \mathcal{M}_E |=\{42,64,85,170\}$ for CIFAR and miniImageNet and  $| \mathcal{M}_E |=\{85,128,170,340\}$ for CUB. We also provide the results in tabular form in Appendix~\ref{app_results} (Table~\ref{tableA1}). 

The `Finetune' baselines in these figures correspond to $n_{sc}=0$ case, and  hence, serve as lower bounds to the performance. These figures show that ACC of memory replay methods such as ER-RING and MEGA-I falls sharply and approaches `Finetune' baselines as we reduce the memory size. DER++, which uses both stored labels and logits during replay, performs slightly better than these methods. However, it still exhibits high accuracy drop (up to $\sim 9\%$) when memory size is reduced. In contrast, EPR shows high resilience under extreme memory reduction. For example, accuracy drop is only about $\sim 5\%$ for CIFAR, $\sim 4\%$ for miniImageNet, and $\sim 3\%$ for CUB dataset when memory size is reduced by a factor of 4. Thus, among the memory replay methods for CL, EPR promises to be the best option, especially in the tiny episodic memory regimes.         

\begin{figure*}[t]
\begin{centering}
  \includegraphics[width=0.85\textwidth,keepaspectratio,page=6]{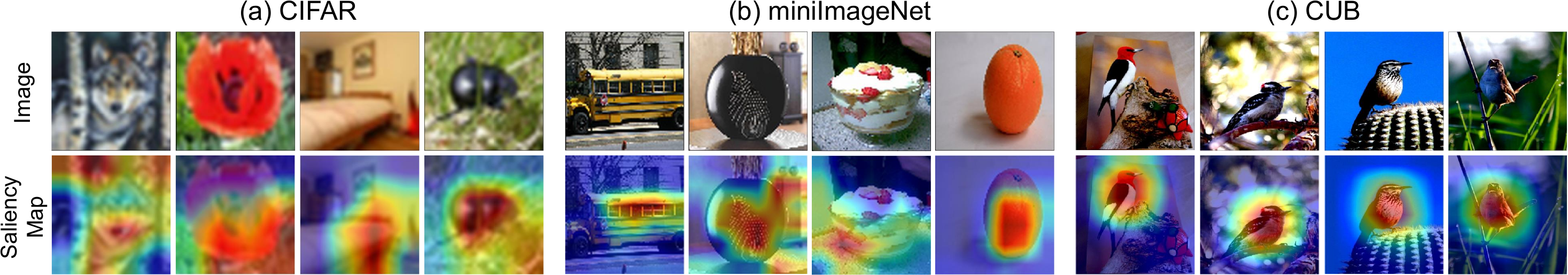}
  \caption{(a) Saliency maps of images from (a) Split CIFAR, (b) Split miniImageNet, (c) Split CUB dataset. CIFAR images have the lowest resolution whereas CUB images have the highest. With increasing image resolution we observe better object localization with Grad-CAM.} 
\label{fig:salmap}
\end{centering}
\vspace{-10pt}
\end{figure*}

\begin{figure}[t]
\begin{centering}
  \includegraphics[width=0.43\textwidth,keepaspectratio,page=5]{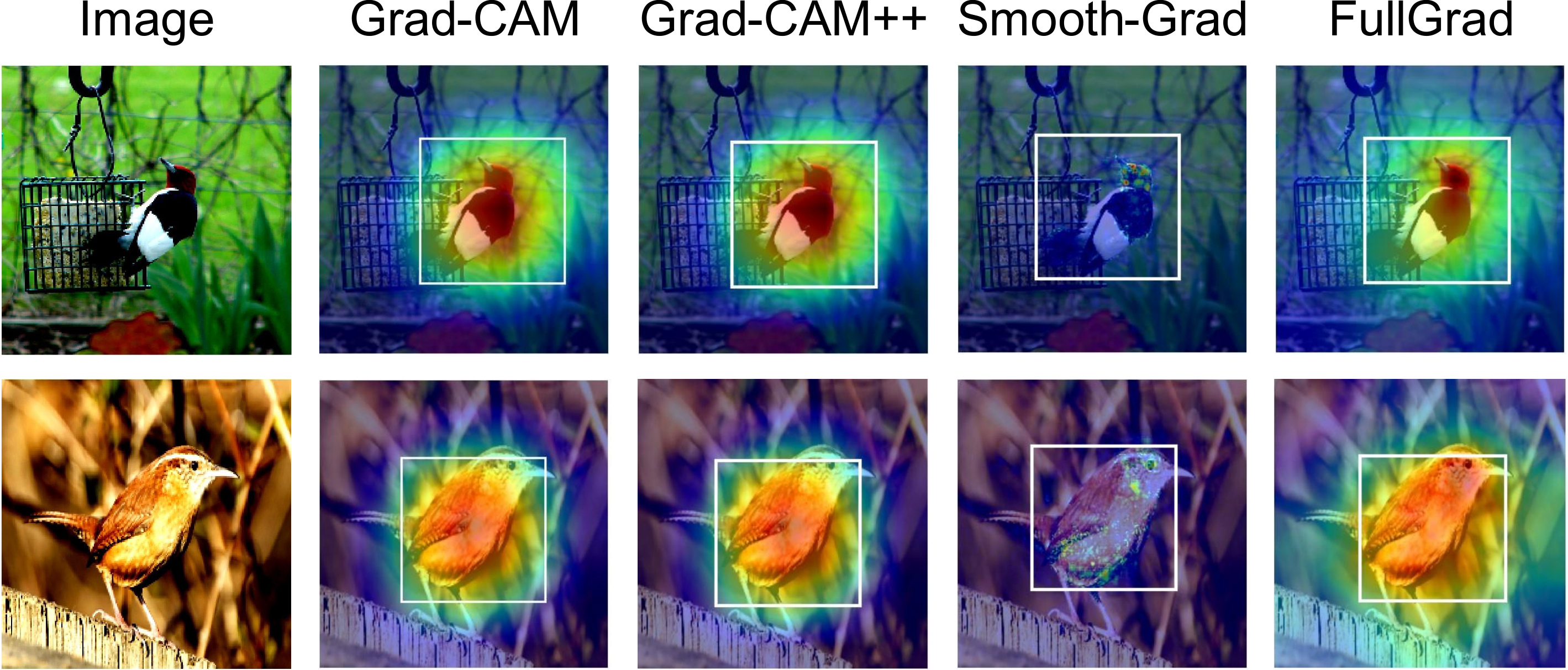}
  \caption{Patch selection from CUB dataset using various saliency methods in EPR.} 
\label{fig:diff_sal}
\end{centering}
\vspace{0pt}
\end{figure}

\textbf{EPF vs. Performance.} In our design, EPF determines how many image patches we can store per class for a given $n_{sc}$. A higher EPF would select smaller patches (Equation~\ref{epf2}), and hence, increase the sample \textit{quantity} (or diversity) per class. However, a large number of memory patches, unlike full images, does not imply better performance from experience replay. In this regard, feature localization quality in the saliency map gives us a better picture about the \textit{quality} of these patches for experience replay. Figure~\ref{fig:salmap} shows the saliency maps of different classes for different datasets from our experiments. For larger-sized and better quality images of CUB dataset, we observe that Grad-CAM localizes the object better within small regions of the images (Figure~\ref{fig:salmap}(c)). This gives us an impression that important part of the image for network decision can be captured with a smaller patch. Thus a higher EPF can be chosen to select a larger number of high quality patches, which would improve the performance of experience replay. In contrast, for smaller-sized and low quality images of CIFAR, we observe that network's decisions are distributed over a large portion of the images (Figure~\ref{fig:salmap}(a)). Thus, a smaller patch (for high EPF) here may not capture enough information to be fully effective in experience replay. Figure~\ref{fig:result_figs}(d) shows the impact of EPF on the performance of our method for different datasets. Since, for $n_{sc}=1$, our methods with EPF~$=1$ is similar to ER-RING, we consider the cases with EPF $\geq2$. For CUB, accuracy of EPR improves as we increase EPF from $2$ to $4$. Beyond that point the accuracy drops which indicates that the memory patches are too small to capture all the relevant information in the given images. For CIFAR, we obtain the best performance for EPF~$=2$ and as we increase EPF we observe drop in accuracy. For miniImageNet, we obtain the best performance for EPF~$=3$. These results support our observations that link the size and quality of the memory patches to the quality of object localization in saliency maps for a given dataset.  

\textbf{Informativeness of Memory Buffer.} Generalization capability of a model trained on the samples from the memory buffer, $\mathcal{M}_E$ can reveal the informativeness of the buffer. Thus, following Buzzega \textit{et al.}~\cite{der}, we compare the informativeness of the EPR buffer with the buffers used in DER++ and ER-RING. For each dataset, we train the corresponding model jointly on all the buffer data from all the tasks. This training does not correspond to CL, rather it mimics the multitasks learning. For EPR buffer, we train the model with zero-padded memory patches.~Figure~\ref{fig:result_figs}(e) shows the average (multitask) accuracy on the test set. For all the datasets, models trained on EPR buffer achieve the highest accuracy (better generalization). Thus, our proposed experience packing method captures richer summaries of underlying data distribution (without any memory increase) compared to the other buffers. This reduces overfitting to the memory buffer, which in turn improves accuracy and reduces forgetting (Table~\ref{table1}, Appendix Table~\ref{tableA1}).    

\textbf{Impact of Saliency Methods on EPR.} To assess the sensitivity of EPR to the choice of saliency algorithms, we also used Grad-CAM++~\cite{grad_cam_pp}, Smooth-Grad~\cite{smoothgrad} and Full-Grad~\cite{fullgrad} as saliency method in EPR. Results are compared in Table~\ref{table3}, where we observe only a marginal variations ($\sim 1\%$) in performance. Corresponding saliency maps (Figure~\ref{fig:diff_sal}) show similar patches are being selected in EPR by these methods. These findings indicate our method can work with a wide variety of saliency methods.     


\renewcommand{\arraystretch}{1.3}
\begin{table}[!h]
\centering
\small
\caption{Impact of various saliency methods on EPR performance.}
\scalebox{0.75}{
\begin{tabular}[t]{@{}lcccccc@{}}
\toprule
 &  \multicolumn{2}{c}{\textbf{CIFAR}} &  \multicolumn{2}{c}{\textbf{miniImageNet}} &  \multicolumn{2}{c}{\textbf{CUB}}\\
\midrule
\textbf{Saliency Method}& {ACC (\%)}  & {BWT}  & {ACC (\%)}  & {BWT} & {ACC (\%)}  & {BWT}\\ \midrule
Grad-CAM~\cite{grad-cam} & \textbf{58.5} & - 0.10  & \textbf{51.9}  & - 0.06 & 72.1 & - 0.02 \\
Grad-CAM++~\cite{grad_cam_pp} & 58.0 & - 0.10  & 51.6  & - 0.07 & 72.4 & - 0.02 \\
Smooth-Grad~\cite{smoothgrad} & 57.5 & - 0.12  & 51.1  & - 0.07 & 72.0 & - 0.02 \\
FullGrad~\cite{fullgrad} & 57.8 & - 0.12   & 51.2  & - 0.07 & \textbf{72.8} & - 0.02 \\
\bottomrule
\end{tabular}
}
\label{table3}
\vspace{0pt}
\end{table}

\textbf{Training Time Analysis.} Finally, following SOTA works, in Figure~\ref{fig:result_figs}(f) we provide training time analysis of the algorithms where algorithm-specific compute overheads are captured in terms of extra time spent. We measured time on a single NVIDIA GeForce GTX 1060 GPU. Compared to the standard replay (ER-RING), EPR only takes up to $\sim30\%$ extra time for training.~This extra time is spent on saliency based patch selection and zero-padding. EPR trades-off this computational overhead by gaining $\sim7\%$ ACC improvement over ER-RING. Compared to other recent works such as DER++ and MEGA-I, EPR trains faster and performs better. HAL did not report training time for the datasets under consideration, and  hence, we could not provide a comparison.~Since, HAL and MER both have meta-optimization steps (higher compute overhead), they are expected to require much larger training time~\cite{hal}.        
\vspace{-5pt}
\section{Conclusions}\label{conc}
\vspace{-5pt}
In this paper, we propose a new experience replay method with small episodic memory for continual learning. Using saliency maps, our method identifies the parts of the input images that are important for model's prediction. We store these patches, instead of full images, in the memory and use them with appropriate zero-padding for replay. Our method thus packs the memory with diverse experiences, and hence captures the past data distribution better without memory increase. Comparison with the SOTA methods on diverse image classification tasks shows that our method is simple, fast, and achieves better accuracy with least amount of forgetting. We believe that the work opens up rich avenues for future research. Firstly, better understanding of the model's decision process and better feature localization with saliency methods would improve the quality of the memory patches and hence improve experience replay. Secondly, new replay techniques for the patches can be explored to reduce the memory overfitting further. Finally, future studies can explore possible applications of our concept in other domains such as in reinforcement learning~\cite{erforrl}.   

\textbf{Acknowledgements.} This work was supported in part by the National Science Foundation, Vannevar Bush Faculty Fellowship, Army Research Office, MURI, and by Center for Brain-Inspired Computing (C-BRIC), one of six centers in JUMP, a SRC program sponsored by DARPA.

{\small
\bibliographystyle{ieee_fullname}
\bibliography{egbib}
}

\clearpage
\setcounter{page}{1}
\setcounter{equation}{0}
\setcounter{figure}{0}
\setcounter{table}{0}
\renewcommand{\theequation}{A.\arabic{equation}}
\renewcommand{\thefigure}{A.\arabic{figure}}
\renewcommand{\thetable}{A.\arabic{table}}
\appendix
\onecolumn

\begin{center}
    \textbf{Saliency Guided Experience Packing for Replay in Continual Learning}
\end{center}

\section*{Appendix}
Section~\ref{App_saliency} describes the steps of saliency map generation using Grad-CAM. Section~\ref{App_dataset} provides the dataset statistics used in different experiments. Pseudo-code of the episodic memory update in EPR is given in Section~\ref{App_algorithm}. List of hyperparameters used for the baseline algorithms and our method is given in Section~\ref{App_hyplist}. Additional results are provided in Section~\ref{app_results}. 

\section{Saliency Method : Grad-CAM}\label{App_saliency}
\vspace{-5pt}
Gradient-weighted Class Activation Mapping (\textbf{Grad-CAM})~\cite{grad-cam} is a saliency method that uses gradients to determine the impact of specific feature map activations on a given prediction. Since later layers in the convolutional neural network capture high-level semantics~\cite{highlevel}, taking gradients of a model output with respect to the feature map activations from one such layers identifies which high-level semantics are important for the model prediction. In our analysis, we select this layer and refer to as \textit{target layer}~\cite{rrr}. List of \textit{target layer} for different experiments is given in Table~\ref{tab:targetlayer}.    

\begin{table}[!h]
\centering
\caption{Target layer names in PyTorch package for saliencies generated by different network architectures in Grad-CAM for different datasets.}
\scalebox{0.85}{
\begin{tabular}{@{}lccc@{}}\toprule
\textbf{Dataset} & \textbf{Network} \phantom{a} & \textbf{Target Layer} \\
\midrule
Split CIFAR    & ResNet18 (reduced)  &  \texttt{layer4.1.shortcut}\\
Split miniImageNet    & ResNet18 (reduced)  &  \texttt{layer4.1.shortcut}\\
Split CUB    & ResNet18  &  \texttt{net.layer4.1.conv2}\\
\bottomrule
\end{tabular}
}
\label{tab:targetlayer}
\end{table}
Let's consider the target layer has $M$ feature maps where each feature map, $A^m \in \mathbb{R}^{u\times v}$ is of width $u$ and height $v$. Also consider, for a given image ($I \in \mathbb{R}^{W\times H\times C}$) belonging to class $c$, the pre-softmax score of the image classifier is $y_c$. To obtain the class-discriminative saliency map, Grad-CAM first takes derivative of $y_c$ with respect to each feature map $A^m$. These gradients are then global-average-pooled over $u$ and $v$ to obtain importance weight, $\alpha_m^c$ for each feature map:
    \vspace{-10pt}
\begin{equation}
    \alpha_m^c = \frac{1}{uv} \sum_{i=1}^u\sum_{j=1}^v \frac{\partial y_c}{\partial A_{ij}^m}, 
\end{equation}
where $A^m_{ij}$ denotes location $(i,j)$ in the feature map $A^m$. Next, these weights are used for computing linear combination of the feature map activations, which is then followed by ReLU to obtain the localization map :
    \vspace{-10pt}
\begin{equation}
    L^c_{Grad-CAM} = \textnormal{ReLU}~(\sum_{m=1}^M \alpha_m^cA^m)
\end{equation}
This map is of the same size ($u\times v$) of $A^m$. Finally, saliency map, $I_{sm} \in \mathbb{R}^{W\times H}$ is generated by upsampling $L^c_{Grad-CAM}$ to the input image resolution using bilinear interpolation.
\vspace{-10pt}
\begin{equation}
    I_{sm} = \texttt{Upsample}~(L^c_{Grad-CAM})
\end{equation}


\setcounter{equation}{0}
\setcounter{figure}{0}
\setcounter{table}{0}
\renewcommand{\theequation}{B.\arabic{equation}}
\renewcommand{\thefigure}{B.\arabic{figure}}
\renewcommand{\thetable}{B.\arabic{table}}

\section{Dataset Statistics}\label{App_dataset}
\renewcommand{\arraystretch}{1.1}
\begin{table*}[!h]
\centering
\caption{Statistics of the CIFAR-100, miniImageNet and CUB datasets used in task-incremental learning experiments.}
\vspace{5pt}
\scalebox{0.85}{
\begin{tabular}{@{}lccccc@{}}\toprule\label{tab_app1}
&  \multicolumn{1}{c}{\textbf{Split CIFAR}} & \phantom{a}& \multicolumn{1}{c}{\textbf{Split miniImageNet}} & \phantom{a}& \multicolumn{1}{c}{\textbf{Split CUB}}\\
\midrule
num. of tasks    & 20 &&   20  && 20   \\
input size ($W\times H\times C$)  & $32\times 32\times 3$ && $84\times 84\times 3$ && $224\times 224\times 3$\\
num. of classes/task  & 5 && 5 && 10  \\
num. of training samples/tasks  &2,500&& 2,500 &&300\\
num. of test samples/tasks & 500&& 500  &&290\\
\bottomrule
\end{tabular}
}
\label{tab:data_stat}
\end{table*}

\clearpage
\section{Memory Update Algorithm}\label{App_algorithm}
\begin{algorithm}
\small
  \caption{Procedure for saliency guided episodic memory update in EPR}\label{epr_algo_app}
  \begin{algorithmic}[1]
    \Procedure{UPDATEMEMORY}{$\mathcal{M}_E,\mathcal{M}_T,f_\theta,\textnormal{EPF},W_p$}
    \State $\texttt{XAI}$: Procedure for saliency map generation; $S_{sm}$: stride; $t^k$: task-ID  
    \State Initialize: $\textbf{I}_p\gets [\;]$; $\textbf{c}\gets[\;]$;  $\textbf{x}_{cord}\gets [\;]$; $\textbf{y}_{cord}\gets [\;]$;  $\textbf{P}_{pred}\gets[\;]$ \Comment{Initialize for memory selection}
    \For {$(I,k,c)\sim \mathcal{M}_T$ } \Comment{Sample one example at a time without replacement from $\mathcal{M}_T$ }
        \State $I_{sm}\gets \texttt{XAI}(f_\theta,I,c)$ \Comment{generate saliency map using Equation~\ref{eq1}}
        \State $x_{cord},y_{cord}\gets \texttt{average-pool}(I_{sm}, W_p, S_{sm})$ \Comment{get corner coordinates of the most salient region in input,$I$}
        \State $I_p\gets I(x_{cord}:x_{cord}+W_p, y_{cord}:y_{cord}+W_p )$ \Comment{get patch from Equation~\ref{eq:epr_eq4}}
        \State $I_p^{'}\gets \texttt{Zero-pad}(I_p,x_{cord},y_{cord})$
        \State $pred\gets f_\theta(I_p^{'})$ \Comment{check model prediction after zero-padding}
        \State $\textbf{I}_p\gets[\textbf{I}_p,I_p]$ \Comment{add patch}
        \State $\textbf{P}_{pred}\gets[\textbf{P}_{pred},pred]$ \Comment{add prediction}
        \State $\textbf{c}\gets[\textbf{c},c]$ \Comment{add class label}
        \State $\textbf{x}_{cord}\gets[\textbf{x}_{cord},x_{cord}]$ \Comment{add $x_{cord}$}
        \State $\textbf{y}_{cord}\gets[\textbf{y}_{cord},y_{cord}]$ \Comment{add $y_{cord}$}
     \EndFor
     \State $(\textbf{I}_p,\textbf{c},\textbf{x}_{cord},\textbf{y}_{cord})\gets \texttt{select-patches} (\textbf{I}_p,\textbf{c},\textbf{x}_{cord},\textbf{y}_{cord},\textbf{P}_{pred}, \textnormal{EPF} )$ \Comment{see section~\ref{sec:res}: memory patch selection}
    \State $t^k\gets k$
     \State $\mathcal{B}_{\mathcal{M}_E}\gets (\textbf{I}_p,t^k,\textbf{c})$
     \State $\mathcal{M}_E\gets \mathcal{M}_E\cup\{(\mathcal{B}_{\mathcal{M}_E},\textbf{x}_{cord},\textbf{y}_{cord})\}$ \Comment{update episodic memory}
     \State \textbf{return} $\mathcal{M}_E$ 
     \EndProcedure

  \end{algorithmic}
\end{algorithm}


\clearpage
\setcounter{equation}{0}
\setcounter{figure}{0}
\setcounter{table}{0}
\renewcommand{\theequation}{D.\arabic{equation}}
\renewcommand{\thefigure}{D.\arabic{figure}}
\renewcommand{\thetable}{D.\arabic{table}}

\section{List of Hyperparameters}\label{App_hyplist}
\vspace{-5pt}
List of hyperparameters used for both baseline methods and our approach is provided in Table~\ref{tab:hyper}. EPF values used in different experiments in our method are given in Table~\ref{table_epf}.
\renewcommand{\arraystretch}{1.2}
\begin{table*}[!h]
\centering
\caption{Hyperparameters grid considered for the baselines and our approach. The best values are given in parentheses. Here, `$lr$' represents learning rate. In the table, we represent Split CIFAR as `cifar', Split miniImageNet as `minImg' and Split CUB as `cub'. EPF is experience packing factor and $\mathcal{M}_T$ is the temporary ring buffer in EPR.}
\scalebox{0.85}{
\begin{tabular}{@{}lcl@{}}\toprule\label{tab_app3}
\textbf{Methods}& \phantom{a} & \textbf{Hyperparameters} \\
\midrule
Finetune    & & $lr$ :~0.003, 0.01, 0.03 (cifar, minImg, cub), 0.1, 0.3, 1.0 \\
\midrule
EWC    & & $lr$ : ~0.003, 0.01, 0.03 (cifar, minImg, cub), 0.1, 0.3, 1.0    \\
       & & regularization, $\lambda$ : 0.1, 1, 10 (cifar, minImg, cub), 100, 1000 \\
\midrule
RRR    & & $lr$ :~0.003, 0.01 (cub), 0.03, 0.1, 0.3, 1.0   \\
       & & regularization : 10, 100 (cub), 1000 \\
\midrule
A-GEM    & & $lr$ :~0.003, 0.01, 0.03 (cifar, minImg, cub), 0.1, 0.3, 1.0   \\
\midrule
MER    & & $lr$ :~0.003, 0.01, 0.03 (cifar, minImg), 0.1 (cub), 0.3, 1.0   \\
       & & with in batch meta-learning rate, $\gamma$ :~0.01, 0.03, 0.1 (cifar, minImg, cub), 0.3, 1.0  \\
       & & current batch learning rate multiplier, $s$ :~1, 2, 5 (cifar, minImg, cub), 10  \\
\midrule
MEGA-I   & & $lr$ :~0.003, 0.01, 0.03 (cifar, minImg, cub), 0.1, 0.3, 1.0   \\
       & & sensitivity parameter, $\epsilon$ :~$1e^{-5}$, $1e^{-4}$, 0.001, 0.01 (cifar, minImg, cub), 0.1  \\
\midrule
DER++   & & $lr$ :~0.003, 0.01, 0.03 (minImg, cub), 0.1 (cifar), 0.3, 1.0   \\
       & & regularization $\alpha$ :~0.1 (minImg), 0.2 (cifar), 0.5 (cub), 1.0  \\
       & & regularization, $\beta$ :~0.5 (cifar, minImg, cub), 1.0  \\
\midrule
ASER    & & $lr$ :~0.003, 0.01, 0.03 (cub), 0.1 (cifar, minImg), 0.3, 1.0 \\
& & $K$ : 3 (cifar, minImg, cub); $N_c$ : 100 (cifar, miniImg), 150 (cub), 250 \\
\midrule
ER-Reservoir    & & $lr$ :~0.003, 0.01, 0.03 (cub), 0.1 (cifar, minImg), 0.3, 1.0 \\
\midrule
ER-RING    & & $lr$ :~0.003, 0.01, 0.03 (cifar, minImg, cub), 0.1, 0.3, 1.0 \\
\midrule
HAL    & & $lr$ :~0.003, 0.01, 0.03 (cifar, minImg), 0.1, 0.3, 1.0   \\
       & & regularization, $\lambda$ : 0.01, 0.03, 0.1, 0.3 (minImg), 1 (cifar), 3, 10 \\
       & & mean embedding strength, $\gamma$ :~0.01, 0.03, 0.1 (cifar, minImg), 0.3, 1, 3, 10  \\
       & & decay rate, $\beta$ : 0.5 (cifar, minImg)  \\
       & & gradient steps on anchors, $k$ : 100 (cifar, minImg)  \\
       
\midrule
Multitask & & $lr$ :~0.003, 0.01, 0.03 (cifar, minImg, cub), 0.1, 0.3, 1.0 \\
\midrule
EPR (ours) & & $lr$ (task-incremental) :~0.01, 0.03 (cub), 0.05 (minImg), 0.1 (cifar), 0.3, 1.0 \\
 & & $lr$ (class-incremental) :~0.01, 0.05 (cifar, minImg), 0.1  \\
& & examples per class temporarily stored in $\mathcal{M}_T$ : $\gamma~\times$ EPF; $\gamma$ : 2(cub), 5 (cifar,minImg) \\
& & stride, $S_{sm}$ : 1 (cifar, minImg), 2, 3 (cub)  \\
\bottomrule
\end{tabular}
}
\label{tab:hyper}
\end{table*}



\renewcommand{\arraystretch}{1.1}
\begin{table*}[!h]
\caption{Experience Packing Factor (EPF) for different $n_{sc}$ used in our (a) task-incremental learning and (b) class-incremental learning experiments. Input image width, $W$ for CIFAR, miniImageNet and CUB dataset are $32,84$ and $224$ respectively. For given $n_{sc}$, EPF and $W$, corresponding memory patch sizes ($W_p$) are also given in the table. }
\hspace{1cm}
\vspace{-25pt}
\begin{minipage}[t]{0.1\linewidth}
\centering
\scalebox{0.7}{
\begin{tabular}[t]{@{}ccccccccc@{}}
&  \multicolumn{7}{c}{} \\
&  \multicolumn{7}{c}{\textbf{(a)}} \\
\toprule
&  \multicolumn{2}{c}{\textbf{Split CIFAR}} & & \multicolumn{2}{c}{\textbf{Split miniImageNet}} & & \multicolumn{2}{c}{\enspace\enspace\textbf{Split CUB}\enspace\enspace}\\
\midrule
\enspace\enspace\textbf{$n_{sc}$}\enspace\enspace& {EPF}  & {$W_p$}  && \enspace\enspace{EPF}\enspace  & {$W_p$} && {EPF}  & \enspace{$W_p$}\enspace\\ \midrule
2 & 3 &  26 &&  5 & 53 && 7 & 119  \\
1 & 2 & 22  &&  3 & 48 && 4 & 112  \\
0.75 & 1 & 27   && 2 & 51 && 3 & 112 \\
0.5 & 1 & 22   &&  2 & 42  &&2 & 112 \\

\bottomrule
\end{tabular}
}
\end{minipage}
\hspace{0.5cm}
\begin{minipage}[t]{1.05\linewidth}
\centering
\scalebox{0.8}{
\begin{tabular}[t]{@{}cccccccccc@{}}
&  \multicolumn{6}{c}{} \\
&  \multicolumn{5}{c}{\textbf{(b)}} \\
\toprule
& &  \multicolumn{2}{c}{\textbf{CIFAR-100}} & & \multicolumn{2}{c}{\textbf{miniImageNet}} \\
& &  \multicolumn{2}{c}{(20 Tasks)} & & \multicolumn{2}{c}{(10 Tasks)} \\
\midrule
\enspace\textbf{$\vert M_E \vert$}\enspace & \enspace\textbf{$n_{sc}$}\enspace& {EPF}  & {$W_p$}  && \enspace\enspace{EPF}\enspace  & {$W_p$} \\ \midrule
2k &20 & 25 & 28 &&  25 & 75  \\
1k &10 & 13 & 28  && 13 & 73   \\
\bottomrule
\end{tabular}
}
\end{minipage}
\label{table_epf}
\end{table*}

\clearpage
\setcounter{equation}{0}
\setcounter{figure}{0}
\setcounter{table}{0}
\renewcommand{\theequation}{E.\arabic{equation}}
\renewcommand{\thefigure}{E.\arabic{figure}}
\renewcommand{\thetable}{E.\arabic{table}}

\section{Additional Results}\label{app_results}

\renewcommand{\arraystretch}{1.2}
\begin{table*}[!h]
\centering
\caption{Performance comparison of different experience replay methods for different memory sizes in task-incremental learning setup. Number of memory slots per class, $n_{sc}$=$\{0.5,0.75\}$ refers to memory size, $| \mathcal{M}_E |$=$\{42,64\}$ for CIFAR and miniImageNet, and $| \mathcal{M}_E |$=$\{85,128\}$ for CUB. Average and standard deviations are computed over $5$ runs for different random seeds.}
\vspace{5pt}
\scalebox{0.8}{
\begin{tabular}[!h]{@{}clccccccccc@{}}
\toprule
&\multicolumn{1}{c}{}
&\multicolumn{2}{c}{\textbf{Split CIFAR}} 
&\multicolumn{3}{c}{\textbf{Split miniImageNet}} 
&\multicolumn{3}{c}{\textbf{Split CUB}} \\

\midrule
\textbf{$n_{sc}$}&\textbf{Methods}& {ACC (\%)}  & {BWT}  && {ACC (\%)}  & {BWT} && {ACC (\%)}  & {BWT}\\ \midrule
-&Finetune    &  42.9 $\pm$ 2.07 & - 0.25 $\pm$ 0.03 &&  34.7 $\pm$ 2.69 & - 0.26 $\pm$ 0.03  && 55.7 $\pm$ 2.22 & - 0.13 $\pm$ 0.03 \\
\midrule

&MEGA-I & 48.9 $\pm$ 1.68 & - 0.21 $\pm$ 0.01  && 43.8 $\pm$ 1.58  & - 0.14 $\pm$ 0.01 && 61.5 $\pm$ 2.08 & - 0.08 $\pm$ 0.01\\
0.75&DER++ &  50.0 $\pm$ 1.81 & - 0.19 $\pm$ 0.02 &&  47.2 $\pm$ 1.54& - 0.12 $\pm$ 0.01 &&  64.8 $\pm$ 1.61 & - 0.06 $\pm$ 0.01\\
&ER-RING &  50.4 $\pm$ 0.85 & - 0.21 $\pm$ 0.02 &&  44.9 $\pm$ 1.49 & - 0.14 $\pm$ 0.02 &&  64.0 $\pm$ 1.29 & - 0.05 $\pm$ 0.01\\
\rowcolor{Gray}
&\textbf{EPR (Ours)} & \textbf{56.8 $\pm$ 1.59} & - \textbf{0.12 $\pm$ 0.02}  && \textbf{51.1 $\pm$ 1.47}  & - \textbf{0.06 $\pm$ 0.01} && \textbf{70.7 $\pm$ 0.72} & - \textbf{0.03 $\pm$ 0.01}\\
\midrule

&MEGA-I & 43.7 $\pm$ 1.26 & - 0.26 $\pm$ 0.02  && 39.6 $\pm$ 2.35  & - 0.18 $\pm$ 0.02 && 57.7 $\pm$ 0.62 & - 0.11 $\pm$ 0.01\\
0.5&DER++ &  47.5 $\pm$ 1.58 & - 0.21 $\pm$ 0.01 &&  45.6 $\pm$ 0.56& - 0.13 $\pm$ 0.01 &&  62.5 $\pm$ 1.45 & - 0.08 $\pm$ 0.01\\
&ER-RING &  44.6 $\pm$ 0.84 & - 0.27 $\pm$ 0.01 &&  39.1 $\pm$ 1.38 & - 0.20 $\pm$ 0.02 &&  59.2 $\pm$ 0.97 & - 0.10 $\pm$ 0.01\\
\rowcolor{Gray}
&\textbf{EPR (Ours)} & \textbf{55.6 $\pm$ 0.54} & - \textbf{0.13 $\pm$ 0.02}  && \textbf{49.2 $\pm$ 1.20}  & - \textbf{0.07 $\pm$ 0.01} && \textbf{70.3 $\pm$ 0.91} & - \textbf{0.03 $\pm$ 0.01}\\

\bottomrule
\end{tabular}
}

\label{tableA1}
\vspace{-8pt}
\end{table*}

\end{document}